\documentclass{article}
\usepackage{spconf,amsmath,graphicx}
\usepackage{enumitem}
\setlist{nosep, leftmargin=14pt}
\usepackage{url}

\usepackage{float}

\title{Feature-Spectral Fragility in Segmentation: Dataset Dependence, Architecture-Specific Localization, and Spectral Correlates}

\name{Subhash Kashyap}
\address{Department of Computer Science and Engineering\\
National Institute of Technology Rourkela\\
Odisha, India}

\begin{document}

\maketitle

\begin{abstract}
Robustness of segmentation models is commonly assessed through
input-domain perturbations, while dependence on frequency content within
learned feature representations remains less understood. We probe this
dependence using targeted post-training low-pass interventions on
internal representations of three segmentation architectures, ResNet50-UNet
(CNN), VM-UNet (SSM), and Swin-UNETR (Transformer), across CVC-ClinicDB
and ISIC2018, with headline evaluations performed on untouched held-out
test sets. At cutoff $\rho=0.25$, feature-domain low-pass filtering
causes severe degradation on CVC: Dice drops by $100\%$, $73.2\%$, and
$30.9\%$ for CNN, SSM, and Transformer, respectively, compared with
$9.4\%$, $10.3\%$, and $0.6\%$ on ISIC. The cross-dataset difference
is statistically significant for every architecture. Single-stage
interventions further show that the most sensitive stage depends on
architecture and dataset: the CNN peaks at the 2nd to 3rd encoder
block, whereas the SSM peaks at the 1st to 2nd encoder stage. Native feature-domain spectral measurements show an inverse
association between high-frequency energy and fragility on CVC; the
relationship is only partial on ISIC and is therefore treated as a
candidate correlate rather than a proven mechanism. Finally, Fourier
augmentation improves robustness to input-space low-pass filtering
but leaves feature-domain degradation essentially unchanged. These
results show that feature-spectral robustness is strongly
dataset-dependent, architecture-specific, and distinct from
input-domain spectral robustness.
\end{abstract}


\section{Introduction}
\label{sec:intro}

Robustness of medical image segmentation models is commonly assessed with
\emph{input-domain} perturbations such as noise, blur, or domain shift
\cite{drenkow2022robustness}. Such tests measure whether a trained model
remains reliable when the observed input changes, but they do not
directly test whether predictions depend on specific properties of the
\emph{learned representation}. A feature-domain intervention that
selectively modifies the frequency content of an internal representation
therefore provides a direct probe of representation dependence
\cite{geirhos2020shortcut}.

Frequency bias in vision models has primarily been studied as a question
of which input or representation statistics models exploit
\cite{geirhos2020shortcut}. Less is known about three related questions
in segmentation: \emph{how strongly} predictions depend on feature-domain
frequency content, \emph{where} that dependence is localized within a
network, and whether native feature spectra provide a measurable correlate
of the observed fragility. These questions span substantially different
segmentation architectures, including convolutional, state-space, and
transformer-based models
\cite{gu2023mamba,ruan2024vmunet,hatamizadeh2022swinunetr}.

We study feature-spectral fragility through a common post-training
intervention framework across three segmentation architectures and two
datasets. We quantify whole-network feature-domain fragility, localize
sensitivity through single-stage interventions, and measure native
feature-domain spectral energy as a candidate correlate of fragility.
All headline evaluations use untouched held-out test splits. We find:
(i) feature-spectral fragility is dramatically larger on CVC than ISIC
for every evaluated architecture; (ii) the most sensitive stage varies with architecture and dataset, with
CNN effects peaking at the 2nd to 3rd encoder block and SSM effects at
the 1st to 2nd encoder stage; and (iii) native
high-frequency energy shows an inverse association with fragility on CVC,
although this relationship is not universal across datasets. Finally,
Fourier augmentation improves input-domain spectral robustness without
removing the corresponding feature-domain fragility.

\section{Methods}
\label{sec:methods}

\subsection{Datasets and splits}
\label{ssec:data}

CVC-ClinicDB (612 polyp frames) and ISIC2018 (2{,}594 lesion images)
\cite{codella2019isic,bernal2015cvc}. Deterministic splits (seed 42)
produce 489/61/62 and 2{,}075/259/260 train/validation/test images for
CVC and ISIC, respectively. Validation data were used for model
selection; test splits were untouched until final evaluation. All
headline results use the held-out test sets ($n=62$ for CVC and
$n=260$ for ISIC).

\subsection{Architectures}
\label{ssec:arch}

\textbf{CNN:} ResNet50-UNet. The models are trained with
BCEWithLogits loss, Adam with learning rate $10^{-4}$, cosine annealing
for 100 epochs, 256-pixel inputs, batch size 8, and seed 42. Input
normalization is ImageNet-style for ISIC and $[0,1]$ scaling for CVC.

\textbf{SSM:} VM-UNet with 30 VSSBlocks and encoder depths
$[2,2,9,2]$. The CVC and ISIC models are evaluated using their
respective trained checkpoints under the same feature-domain
intervention protocol described below.

\textbf{Transformer:} Swin-UNETR, trained with BCEWithLogits loss,
Adam with learning rate $10^{-4}$, cosine annealing for 100 epochs,
256-pixel inputs, batch size 8, and seed 42.

\subsection{Interventions}
\label{ssec:interv}

\textbf{Whole-network LP:} Forward hooks are placed on semantic feature
outputs and an ideal circular low-pass mask is applied in the 2-D DFT,
setting frequencies above normalized radial cutoff
$\rho \in \{0.10,\ldots,0.40\}$ to zero before inverse transformation.
The headline comparison uses $\rho=0.25$. All interventions are
post-training; model parameters are unchanged.

\textbf{Single-stage localization:} To identify where feature-spectral
sensitivity is concentrated, the same mask is applied to exactly one
semantic stage at a time, with all other representations left unchanged.
CNN localization is evaluated at native block granularity. For VM-UNet,
the 30 VSSBlocks are grouped into its eight native encoder/decoder
stage-groups and the final VSSBlock of each group is intervened on.
Because a CNN block and an SSM stage-group cover different fractions of
their respective networks, absolute $\Delta$Dice magnitudes are compared
only within an architecture; cross-architecture comparison is therefore
restricted to the relative depth profile.

\textbf{Native spectral energy:} With no intervention, we compute the
2-D FFT power spectrum of each stage's clean output and report the
fraction of energy in low ($\leq25\%$ Nyquist), mid (25--60\%), and
high ($>60\%$) radial frequency bands at four spatial resolutions per
architecture. CNN/SSM resolutions are $\{64,32,16,8\}$; Transformer
resolutions are $\{128,64,32,16\}$. Cross-architecture comparison is
therefore made only at matched spatial resolution, with the primary
comparison at $16\times16$.

\subsection{Metrics and statistics}
\label{ssec:metrics}

We report pooled and per-image Dice. Paired Wilcoxon signed-rank tests
are applied to per-image Dice for intervention effects. We use
10{,}000-resample bootstrap 95\% confidence intervals, including a
two-sample bootstrap for the CVC-versus-ISIC difference in mean
Dice drop per architecture. We define
\[
\Delta\mathrm{Dice}
=
\mathrm{Dice}_{\mathrm{intervened}}
-
\mathrm{Dice}_{\mathrm{clean}},
\]
so negative values indicate degradation. For the cross-dataset
comparison we use the Dice drop, $-\Delta\mathrm{Dice}$, so larger
positive values indicate greater degradation. All training experiments use
a single seed (42).

\section{Results}
\label{sec:results}

\subsection{Whole-network effect and cross-dataset dependence}
\label{ssec:effect}

At $\rho=0.25$, applying the feature-domain low-pass intervention causes
severe degradation on CVC. Dice falls from 0.960 to 0.000 for the CNN
($-100\%$), while the SSM and Transformer degrade by $-73.2\%$ and
$-30.9\%$, respectively. On ISIC, degradation is much smaller
($-9.4\%$, $-10.3\%$, and $-0.6\%$); the CNN and SSM are
nearly tied, with SSM slightly ahead (SSM $>$ CNN $>$
Transformer), whereas on CVC the ordering is
CNN $>$ SSM $>$ Transformer.

The within-dataset intervention effects are significant at the
per-image level. More importantly, the
cross-dataset difference is itself statistically resolvable:the 95\%
bootstrap confidence intervals for the difference in mean Dice drop
between CVC and ISIC are $[0.799,0.841]$ for CNN,
$[0.543,0.668]$ for SSM, and $[0.237,0.355]$ for the Transformer.

\textbf{Input-feature dissociation.}
Input-domain low-pass filtering of equal nominal strength is substantially
less destructive on CVC, reducing Dice by $20.0\%$, $6.2\%$, and $0.8\%$
for the CNN, SSM, and Transformer, respectively. Fourier augmentation
during SSM training further reduces input-domain degradation from
$-6.2\%$ to $-1.1\%$, yet feature-domain degradation remains
$-73.2\%$. Thus, robustness to spectral perturbations at the input does
not automatically transfer to the learned feature representation.

\subsection{Localization: where does the dependence live?}
\label{ssec:localization}

\begin{figure*}[t]
\centering
\includegraphics[width=0.92\textwidth]{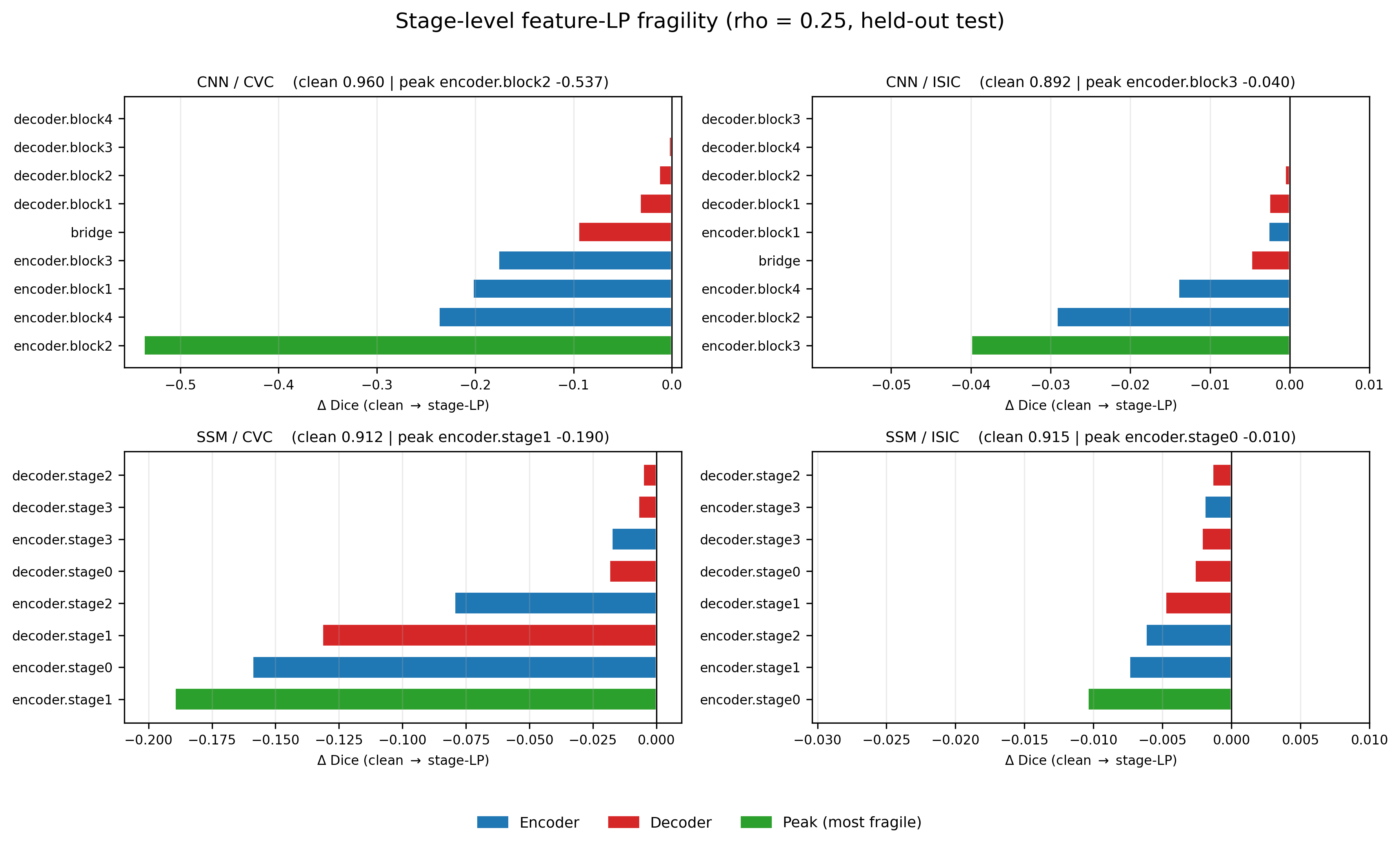}
\caption{Single-stage $\Delta$Dice localization on held-out data.
Peak sensitivity lies at the 2nd to 3rd encoder block for the CNN and
the 1st to 2nd encoder stage for the SSM. Absolute cross-architecture
magnitudes are not directly comparable because intervention granularity
differs.}
\label{fig:localization}
\end{figure*}

Single-stage intervention reveals that feature-spectral sensitivity is
not distributed uniformly through the network
(Fig.~\ref{fig:localization}, Table~\ref{tab:t3}). The CNN shows its
strongest effect at \texttt{encoder.block2} on CVC
($\Delta$Dice $=-0.537$) and at \texttt{encoder.block3} on ISIC
($-0.040$), the 2nd and 3rd of four encoder blocks. The SSM peaks at
\texttt{encoder.stage1} on CVC ($-0.190$) and at
\texttt{encoder.stage0} on ISIC ($-0.0104$), the 2nd and 1st of four
encoder stages.

On CVC, where the effects are large, both architectures peak at the
2nd encoder stage. The SSM peak on ISIC is very small and should not be
over-read. We conclude only that peak sensitivity lies in the early to
middle encoder, that its exact location depends on architecture and
dataset, and that it does not simply grow with depth. Because stage
granularity differs between CNN and SSM representations, absolute
$\Delta$Dice values are compared only within each architecture.

\subsection{Native high-frequency energy as a candidate correlate}
\label{ssec:explanation}

\begin{figure}[!htb]
\centering
\includegraphics[width=\linewidth]{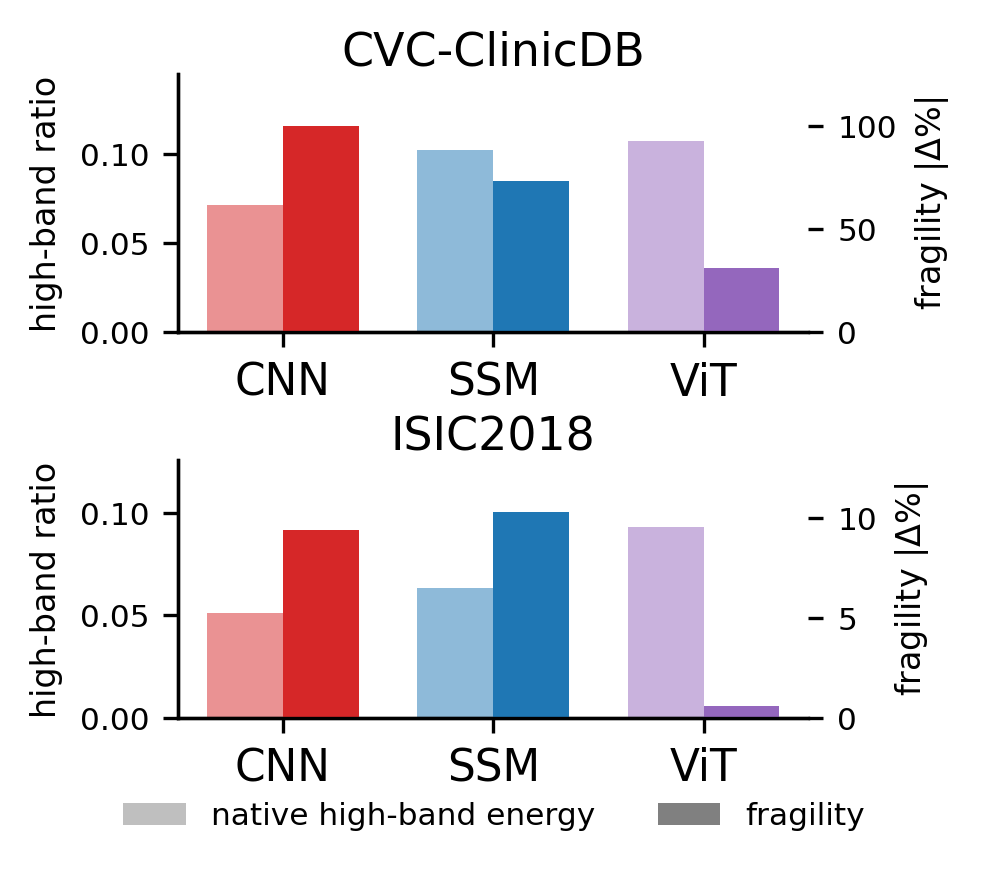}
\caption{Native high-band energy versus feature-domain fragility at matched
$16\times16$ spatial resolution. The inverse ordering is clearest on CVC;
the CNN/SSM pair on ISIC provides a boundary condition rather than a
universal one-to-one relationship.}
\label{fig:explanation}
\end{figure}

At matched $16\times16$ spatial resolution
(Fig.~\ref{fig:explanation}), native high-band energy
on CVC is CNN $0.071 <$ SSM $0.102 <$ Transformer $0.107$, an ordering
that is the exact inverse of the fragility ordering (CNN most fragile,
Transformer least). This inverse ordering suggests that native spectral composition may be
associated with tolerance to high-frequency removal. However, this
comparison contains only three architecture-level observations and
therefore provides a candidate correlate rather than a demonstrated
mechanism.

The ISIC data provide an important boundary condition. The Transformer
again combines the highest high-band energy (0.093) with the smallest
fragility ($-0.6\%$), whereas the CNN/SSM energy ordering
($0.051$ versus $0.063$) does not reproduce their fragility ordering
($-9.4\%$ versus $-10.3\%$). We retain this mismatch rather than
selectively reporting only supporting cases.

\begin{table}[t]
\centering
\footnotesize
\caption{Feature-domain fragility at $\rho=0.25$ on held-out test sets.}
\label{tab:t2}
\begin{tabular}{lcc}
\hline
Model & CVC ($n=62$) & ISIC ($n=260$) \\
\hline
CNN & $-100.0\%$ & $-9.4\%$ \\
SSM & $-73.2\%^{\dagger}$ & $-10.3\%^{\dagger}$ \\
Transformer & $-30.9\%$ & $-0.6\%$ \\
\hline
\end{tabular}
\\[2pt]
{\footnotesize $^{\dagger}$ISIC SSM uses a different VSSM implementation than CVC (0.906 mean feature similarity).}
\end{table}

\begin{table}[t]
\centering
\footnotesize
\caption{Most-damaging single-stage intervention on held-out data.
Absolute $\Delta$Dice values are compared only within an architecture.}
\label{tab:t3}
\begin{tabular}{lccc}
\hline
Leg & Peak stage & $\Delta$Dice & Encoder position \\
\hline
CNN/CVC & enc.block2 & $-0.537$ & 2nd of 4 \\
CNN/ISIC & enc.block3 & $-0.040$ & 3rd of 4 \\
SSM/CVC & enc.stage1 & $-0.190$ & 2nd of 4   \\
SSM/ISIC & enc.stage0 & $-0.0104$ & 1st of 4  \\
\hline
\end{tabular}
\\[2pt]
{\footnotesize Absolute $\Delta$Dice values are not directly comparable
between CNN blocks and SSM stage-groups because their intervention
scopes differ.}
\end{table}

\begin{figure}[t]
\centering
\includegraphics[width=0.95\linewidth]{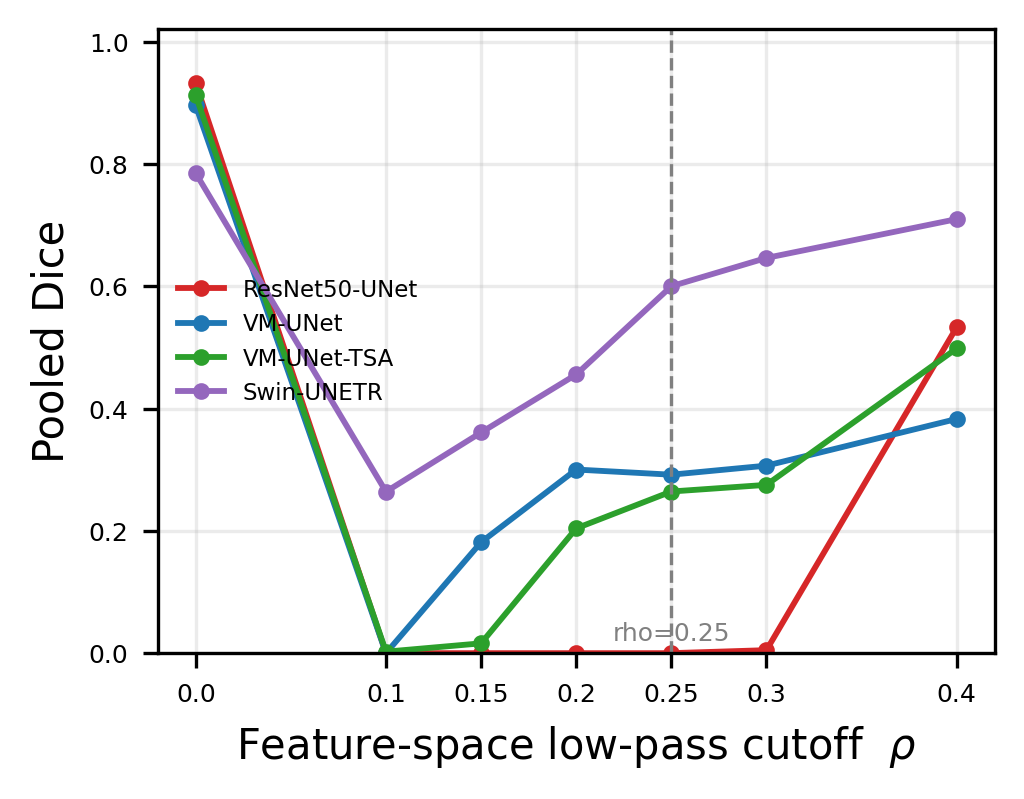}
\caption{CVC feature-domain low-pass dose-response. Clean predictions are
shown at $\rho=0$; the headline intervention uses $\rho=0.25$.}
\label{fig:dose}
\end{figure}

\section{Discussion and Conclusion}
\label{sec:discussion}

Feature-domain spectral fragility is strongly dataset-dependent and is
not uniformly distributed within a network. Feature-domain low-pass
intervention produces severe CVC degradation but substantially smaller
effects on ISIC, with statistically significant cross-dataset differences
for all three architectures. Localization further shows that peak sensitivity lies in the early to
middle encoder, with its exact position depending on architecture and
dataset. Native high-frequency
energy is inversely ordered with fragility on CVC, suggesting a useful
candidate correlate, but the weaker ISIC correspondence shows that this
is not a universal mechanism. Finally, improved robustness to
input-space spectral perturbation does not guarantee robustness of the
learned feature representation.

\textbf{Limitations.}
All models use a single training seed. Localization is performed at each
model's native block/stage granularity, so absolute $\Delta$Dice values
are not compared across architectures. The native spectral-energy
analysis is based on three architecture-level observations and is
therefore hypothesis-generating. The two datasets also differ in image
content and acquisition characteristics, so the observed contrast should
be interpreted as dataset/task dependence rather than attribution to a
single image statistic.

\section{Compliance with Ethical Standards}
This research was conducted retrospectively using fully anonymized,
publicly available human-subject data (CVC-ClinicDB, ISIC2018). No new
data were collected and no identifiable information was accessed. This
study was performed in line with the principles of the Declaration of
Helsinki (1975, revised 2000).

\section{Acknowledgments}
No funding was received. The author has no conflicts of interest to
disclose. AI assistance: Cline helped write parts of the training scripts
and DeepSeek helped write the evaluation scripts. The author reviewed all
code and takes responsibility for the content. Code:
\url{https://github.com/Subkash2206/CausalMamba}


\begin{thebibliography}{99}

\bibitem{geirhos2020shortcut}
R. Geirhos et al., ``Shortcut learning in deep neural networks,''
\emph{Nat. Mach. Intell.}, vol. 2, pp. 665--673, 2020.

\bibitem{drenkow2022robustness}
N. Drenkow et al., ``A systematic review of robustness in deep learning
for computer vision,'' \emph{arXiv:2112.00639}, 2022.

\bibitem{gu2023mamba}
A. Gu and T. Dao, ``Mamba: Linear-time sequence modeling with selective
state spaces,'' \emph{arXiv:2312.00752}, 2023.

\bibitem{ruan2024vmunet}
J. Ruan and S. Xiang, ``VM-UNet: Vision Mamba UNet for medical image
segmentation,'' \emph{arXiv:2402.02491}, 2024.

\bibitem{hatamizadeh2022swinunetr}
A. Hatamizadeh et al., ``Swin UNETR: Swin transformers for semantic
segmentation of brain tumors in MRI images,'' in
\emph{MICCAI BrainLes Workshop}, 2022.

\bibitem{codella2019isic}
N. Codella et al., ``Skin lesion analysis toward melanoma detection
2018: A challenge hosted by the International Skin Imaging Collaboration
(ISIC),'' \emph{arXiv:1902.03368}, 2019.

\bibitem{bernal2015cvc}
J. Bernal et al., ``WM-DOVA maps for accurate polyp highlighting in
colonoscopy: Validation vs.\ saliency maps from physicians,''
\emph{Computerized Medical Imaging and Graphics}, vol. 43,
pp. 99--111, 2015.

\end{thebibliography}
\end{document}